%% file: main.tex
\documentclass[runningheads]{llncs}
\usepackage{graphicx}
\usepackage{comment}
\usepackage{amsmath,amssymb}
\usepackage{color}

\usepackage{algorithm, algorithmic}

\usepackage{booktabs}
\newcommand{\smallsec}[1]{\subsubsection{#1.}}

\newcommand{\pred}{\mathbf{P}}
\newcommand{\gt}{\mathbf{G}}

\def\mymathhyphen{{\hbox{-}}}

\begin{document}

\title{Towards Unique and Informative Captioning of Images}

\titlerunning{ } 
\author{Zeyu Wang\inst{1}\and Berthy Feng\inst{1,2} \and
Karthik Narasimhan\inst{1} \and 
Olga Russakovsky\inst{1}}
\authorrunning{ }
\institute{Princeton University 
 \and
California Institute of Technology \\
}

\maketitle


\begin{abstract}
	
Despite considerable progress, state of the art image captioning models produce generic captions, leaving out important image details. Furthermore, these systems may even misrepresent the image in order to produce a simpler caption consisting of common concepts. In this paper, we first analyze both modern captioning systems and evaluation metrics through empirical experiments to quantify these phenomena. We find that modern captioning systems return higher likelihoods for incorrect distractor sentences compared to ground truth captions, and that evaluation metrics like SPICE can be `topped' using simple captioning systems relying on object detectors. Inspired by these observations, we design a new metric (SPICE-U) by introducing a notion of \textit{uniqueness} over the concepts generated in a caption. We show that SPICE-U is better correlated with human judgements compared to SPICE, and effectively captures notions of diversity and descriptiveness. Finally, we also demonstrate a general technique to improve any existing captioning model -- by using mutual information as a re-ranking objective during decoding. Empirically, this results in more unique and informative captions, and improves three different state-of-the-art models on SPICE-U as well as average score over existing metrics.\footnote{Code is available at \url{https://github.com/princetonvisualai/SPICE-U}.}

\end{abstract}


\section{Introduction}

\begin{figure}[t]
    \centering
    \includegraphics[width=1.0\linewidth]{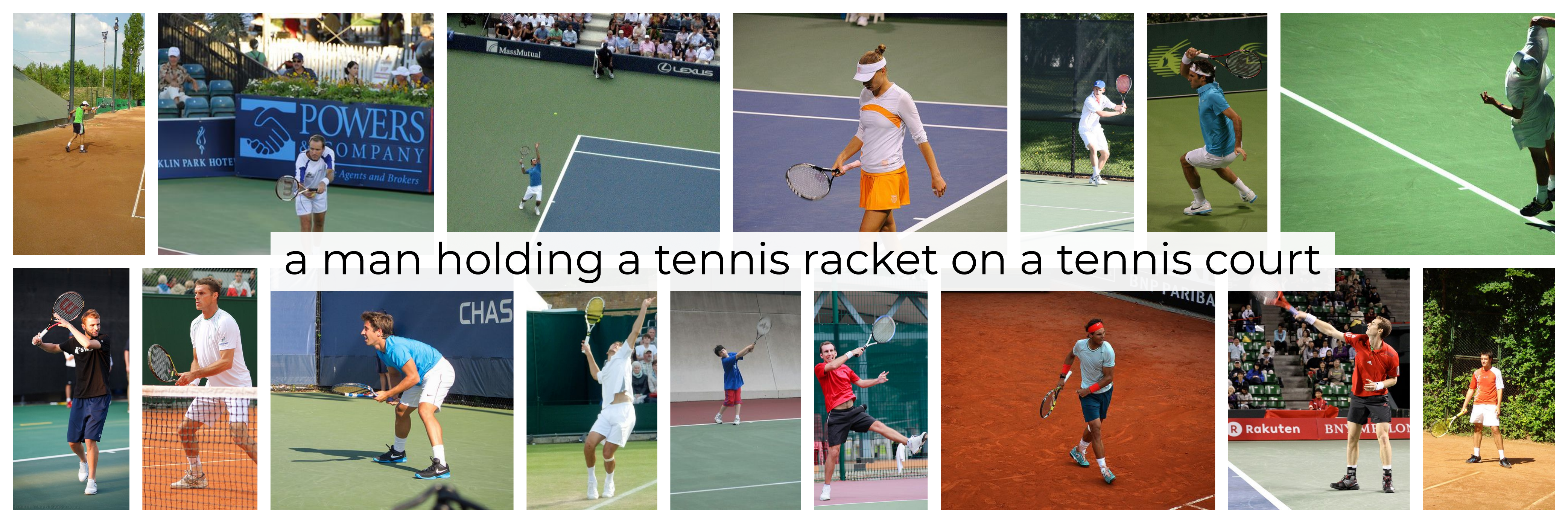}
    \caption{Diverse images from the COCO validation set for which a trained captioning system~\cite{huang_attention_2019} generates the same caption: ``A man holding a tennis racket on a tennis court''. The caption misses important details, such as the action of the person, the type of tennis court, whether there is audience, etc.}
    \label{fig:samecaption}
\end{figure}

Over the last few years, there has been considerable progress in image captioning, with current methods producing fluent captions for a variety of images~\cite{vinyals2015show,xu2015show,you_image_2016,lu_knowing_2017,vinyals2017show,anderson2018bottom,huang_attention_2019}. However, all these systems tend to produce generic captions, re-using a small set of common concepts to describe vastly different images. Consider the example caption in Figure~\ref{fig:samecaption}, produced by a state of the art model~\cite{huang_attention_2019}. Despite obvious differences between the sixteen images, the model produces the same caption, missing several other details specific to certain images and generating incorrect facts about others. A human, on the other hand, would identify unique aspects of each image, such as whether the person is serving, is it a match or a practice, the type of tennis court, the color of the person's shirt, etc. While the inadequacies of the captioning models can be partially attributed to the ``mode collapse'' problem of current techniques and loss functions like cross-entropy, the issue is more fundamental --- defining and benchmarking image captioning adequately remains a challenging task.

To this end, we investigate modern captioning systems in terms of their ability to produce \textit{unique and complete} captions. Specifically, we find that the problem of producing common concepts is deeply ingrained in modern captioning systems. As we demonstrate empirically, one reason for this could be that end-to-end training results in strong language model priors that lead to models preferring more commonly occurring sentences, irrespective of whether they are relevant to the image or not. For instance, we find that state-of-the-art captioning systems~\cite{anderson2018bottom,luo2018discriminability,huang_attention_2019} incorrectly assign higher likelihoods to irrelevant common captions compared to even ground truth captions paired with a particular image.  Furthermore, we also show that this is not just a problem with the captioning models -- existing evaluation metrics frequently fail to reward diversity and uniqueness in captions, in fact \emph{preferring} simple automatically generated captions to more descriptive captions produced by human annotators.  

In this paper, we take a step towards quantitatively characterizing these deficiencies by proposing a new measure which captures the ability of a caption to uniquely identify an image. We convert a caption into a set of objects, attributes, and relations. For each such concept, we compute its uniqueness as a function of the global number of images containing the concept. This is then aggregated over concepts to compute the uniqueness of the overall caption. This uniqueness metric is orthogonal to standard measures like precision and recall, and allows us to combine them using a harmonic mean to define a new metric, SPICE-U. We empirically demonstrate that this metric correlates better with human judgements than the commonly-used SPICE~\cite{spice2016} metric.

Next, we propose techniques to improve current captioning systems at producing unique, more meaningful captions. We employ the strategy of re-ranking captions during the decoding process by maximizing mutual information between the image and the caption (inspired by a similar line of work in machine translation~\cite{li2016mutual}). Our method achieves an absolute improvement of up to $1.6\%$ in SPICE-U and a relative improvement of up to $2.4\%$ on the average across different metrics. The captions produced are more informative and relevant to the image while not losing out on fluency.

To summarize, the contributions of this paper are:
\begin{itemize}
    \item quantitatively demonstrating limitations of current captioning systems and metrics.
    \item proposing a new metric (SPICE-U) that measures the ability of captions to be unique and descriptive.
    \item investigating new decoding objectives to generate more informative captions.
\end{itemize}


\section{Related work}

\smallsec{Discriminative captioning} A number of recent approaches address the task of producing discriminative captions~\cite{Mao_2016_CVPR,vedantam2017context,liu2018show,luo2018discriminability,liu2019generating}. One such method considers the task of distinguishing a target image from a distractor image using generated captions~\cite{vedantam2017context}. The proposed method balances the objectives of maximizing the probability of seeing the predicted caption conditioned on the target image and minimizing the probability of seeing the predicted caption conditioned on the distractor image. 
Other methods~\cite{luo2018discriminability,liu2018show} incorporate image retrieval into the training process, encouraging the generation of captions that are more likely to be uniquely aligned to the original image than to other images. While these approaches help generate more discriminative captions, they are approximate versions of maximizing mutual information between the image and caption, which we aim to do explicitly. 

\smallsec{Descriptive captioning}
Prior work has also focused on improving the amount of information present in a single generated caption. Dense captioning~\cite{johnson2016densecap} aims to identify all the salient regions in an image and describe each with a caption. Diverse image annotation~\cite{wu2017diverse} focuses on describing as much of the image as possible with a limited number of tags. Entity-aware captioning~\cite{lu2018entity} employs hashtags as additional input. Image paragraph captioning~\cite{krause2017hierarchical,melas2018training} aims to produce more than a single sentence for an image. While these papers do capture some notion of expressiveness, they do not explicitly quantify it or examine trade-offs such as the caption length or uniqueness of generated concepts. 

\smallsec{Diversity and mutual information in captioning}
Another related line of research to this paper is the 
area of diversity-promoting objectives for captioning~\cite{vijayakumar2016diverse,vijayakumar2018diverse,shetty2017speaking,wang2019describing,lindh2018generating}. While the similarity lies in aiming to prevent generic, dull captions, these approaches do not explicitly try to make sure that the information content of the caption matches well with the image. In terms of measuring diversity, some papers propose metrics that use corpus-level statistics to provide coarse judgements~\cite{shetty2017speaking,wang2019describing,lindh2018generating}. For instance, one can measure how distinct a set of different captions are for a single image, or how many different captions a model generates across the entire test set. In contrast, our metric provides measurements for \textit{each image-caption pair} using aggregated corpus-level information.

Using mutual information to re-rank scores has been explored in speech recognition~\cite{bahl1986maximum,povey2002minimum}, machine translation~\cite{li2016diversity,li2016mutual,tu2017neural,kimura2019selecting}, conversational agents~\cite{zhang2018generating}, and multimodal search and retrieval~\cite{datta2017multimodal,henning2018estimating,yao2010co}. Maximizing the mutual information as an objective (during either training or inference) has provided reasonable performance gains on all the above tasks. However, to the best of our knowledge, ours is the first work to explore re-ranking via maximizing mutual information specifically to improve the uniqueness of machine-generated captions.

\smallsec{Image captioning metrics}
The most commonly used metrics for image captioning evaluation are BLEU~\cite{papineni2002bleu}, METEOR~\cite{banerjee2005meteor}, CIDEr~\cite{vedantam2015cider}, and SPICE~\cite{spice2016}. BLEU, METEOR, and CIDEr all rely on $n$-gram matching between the candidate caption and reference captions. BLEU and METEOR are traditionally used in machine translation and thus concerned with syntactical soundness. CIDEr measures the similarity between the candidate caption and ``consensus'' of the set of reference captions, essentially calculating how often $n$-grams in the candidate appear in the reference set. SPICE (Semantic Propositional Image Caption Evaluation) is more concerned with the semantics of a caption. It scores a caption based on how closely it matches the scene graph of the target image, where a scene graph consists of a set of object classes, set of relation types, and set of attribute types. While other metrics capture ``naturalness'' of captions, SPICE correlates better with human judgement by focusing on semantic correctness. Attempts at combining metrics have also been made (e.g. SPIDER~\cite{liu2017improved}). More recent work~\cite{rohrbach2018object} points out existing models often hallucinate objects when generating captions and proposes the CHAIR score to explicitly evaluate this problem. In contrast to the above rule-base metrics, recent work has also proposed learning statistical models to evaluate captioning systems~\cite{cui2018learning,dognin2019adversarial}. While these metrics provide a good measure for the accuracy of a caption, they do not explicitly evaluate how descriptive or informative a caption is. Our metric (SPICE-U) incorporates a new `uniqueness' measure, while also capturing  notions of caption fluency and accuracy through traditional precision and recall components.


\section{Analysis: prevalence and causes of common concepts in captions}
\label{sec:analysis}

Current image captioning systems produce captions that are surprisingly fluent and frequently accurate, but generic and uninformative. We begin by demonstrating that the problem of generating common concept is deeply ingrained in both the current captioning systems and in the evaluation metrics. These two factors are closely related as captioning systems are trained to optimize performance on existing metrics. To analyze the captioning systems in the absence of any pre-defined metrics, we take a look directly at the underlying probability distributions learned by the models; to further demonstrate the brittleness of the metrics we design a simple competing baseline that outperforms state-of-the-art captioning systems on standard metrics (this section). Equipped with this analysis, we then go on to propose a new metric (Section~\ref{sec:uniqueness-metric}) along with a potential technical solution (Section~\ref{sec:methods}) to address the problem.

\subsection{Captioning systems prefer common concepts} Modern captioning systems are trained to maximize the likelihood of generating the correct caption sentence $\mathbf{s}$ conditioned on the image $I$, or $P(\mathbf{s}|I)$. Even though the model is learned jointly and does not neatly decompose, intuitively the probability distribution is influenced by two factors: (1) whether a particular concept appears in the image, and (2) the likelihood that the particular concept would appear in a caption. We run a simple experiment to showcase that the latter language prior plays a surprisingly strong role in modern captioning models, helping to partially explain why the systems frequently resort to returning generic image captions corresponding to common concepts.

\begin{table}[t]
\begin{center}
\caption{Five most common captions in the COCO~\cite{lin2014coco} training set with appearance numbers (based on exact matches over the entire sentence). In Section~\ref{sec:analysis}  we demonstrate that captioning models frequently prefer such distractors to ground truth human captions of the images.}
\label{table:distractors}
\begin{tabular}{l}
\hline
\noalign{\smallskip}
		 a man riding a wave on top of a surfboard (160)\\
         a man flying through the air while riding a skateboard (137)\\
         a man riding skis down a snow covered slope (124)\\
         a man holding a tennis racquet on a tennis court (122)\\
         a large long train on a steel track (116)\\
\hline
\end{tabular}
\end{center}
\end{table}

\begin{figure}[t]
    \centering
    \includegraphics[width=0.75\linewidth]{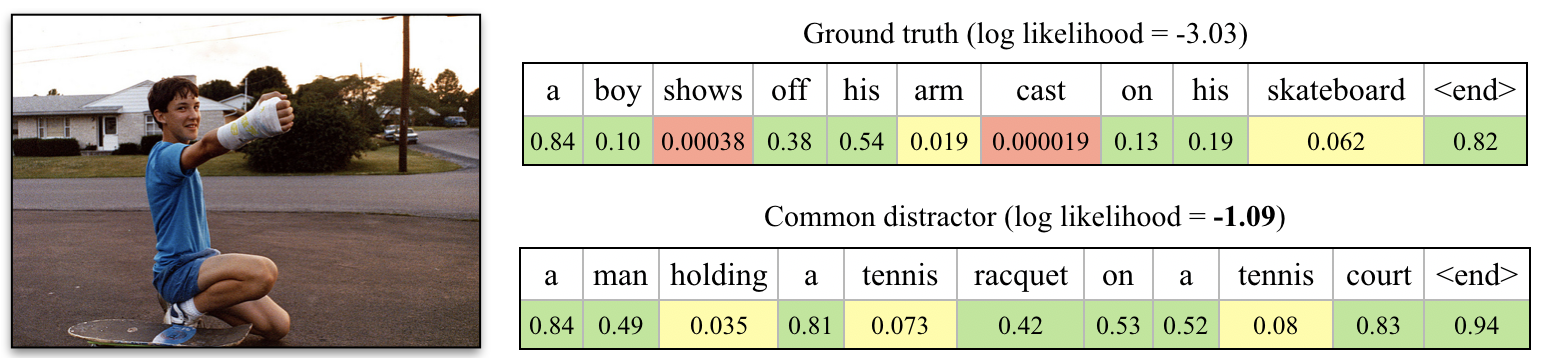}
    \caption{The ground truth caption ``a boy shows off his arm cast on his skateboard'' has much lower mean log likelihood (-3.03) according to the captioning model of~\cite{anderson2018bottom} than a common (on this dataset) but incorrect caption ``a man holding a tennis racquet on a tennis court.'' Numbers under each word $w_k$ correspond to $P(w_k|I,w_{<k})$ of the captioning model, color-coded according to their magnitude. This preference for common captions even at the expense of accuracy is a problem in modern captioning datasets.}
    \label{fig:distractors}
\end{figure}

To do so, we examine the learned probability distribution of the bottom-up top-down attention model \cite{anderson2018bottom} trained on the popular Karpathy split \cite{karpathy2015deep} of the COCO dataset \cite{lin2014coco}. On every validation image, we compare the model's likelihood of the human generated ground truth captions for this image with the model's likelihood corresponding to generic distractor sentences applied to this image. For distractor sentences, we use the five captions that appear most frequently in training set (Table~\ref{table:distractors}). During evaluation, to ensure that these distractor sentences are not correct description of the corresponding image, we use the code from \cite{rohrbach2018object} to only keep the sentences that contain at least one hallucinated object not present in the image. We observe that in an amazing 73\% of the images the model returns a higher likelihood $P(d|I)$ for one of these \emph{wrong} distractor sentences $d$ than its likelihood $P(g|I)$ of one of the ground truth caption, i.e., $\exists d,g: P(d|I) > P(g|I)$.  Figure~\ref{fig:distractors} qualitatively illustrates why this is the case: an incorrect caption associated with common concepts may end up with a higher overall $P(\mathbf{s}|I)$ than a correct caption albeit with rare words, which would receive lower language model scores.

\subsection{Captioning metrics prefer common concepts} We now demonstrate that the problem is not just in the captioning models but also in the metrics used to evaluate those models, such as SPICE~\cite{spice2016}.

\smallsec{Background: SPICE}
While older metrics such as BLEU~\cite{papineni2002bleu}, METEOR~\cite{banerjee2005meteor} and CIDEr~\cite{vedantam2015cider} aim to evaluate both the correctness and the fluency of a caption through n-gram matching, SPICE takes a departure from fluency to focus primarily on caption correctness, i.e., whether the caption reflects visual concepts that are indeed in the image.  Here, a \textit{visual concept} is a concrete thing or abstract notion that can be both localized in an image and described using natural language. For the purposes of evaluation, visual concepts are restricted to objects, their attributes, and their relations~\cite{johnson2015image,spice2016,krishna2017visual}.

Consider an image with a set of visual concepts $\gt$ and a set of predicted visual concepts $\pred$. The accuracy of this description $\pred$ is commonly measured using \emph{precision} and \emph{recall} with regard to the ground truth concepts $\gt$~\cite{spice2016}, where:

 \begin{align}
     &\mathrm{Rec}(\pred;\gt) = \frac{|\pred \cap \gt|}
     {|\gt|}, &&
   \mathrm{Pr}(\pred;\gt)= \frac{|\pred \cap \gt|}
     {|\pred|} \notag  
 \end{align}
 The SPICE metric trades off between them using the harmonic mean:
 \begin{equation}
     \mathrm{SPICE}(\pred;\gt) = \frac{2}{ 1/\mathrm{Rec}(\pred;\gt)+1/\mathrm{Pr}(\pred;\gt)}
     \label{eq:spice}
 \end{equation}
 
We can observe that this metric ignores entirely the uniqueness of concepts and implicitly rewards models which predict common concepts (which are easier to recognize) over rare yet more distinctive concepts. 

\begin{figure}[t]
    \centering
    \includegraphics[width=0.8\linewidth]{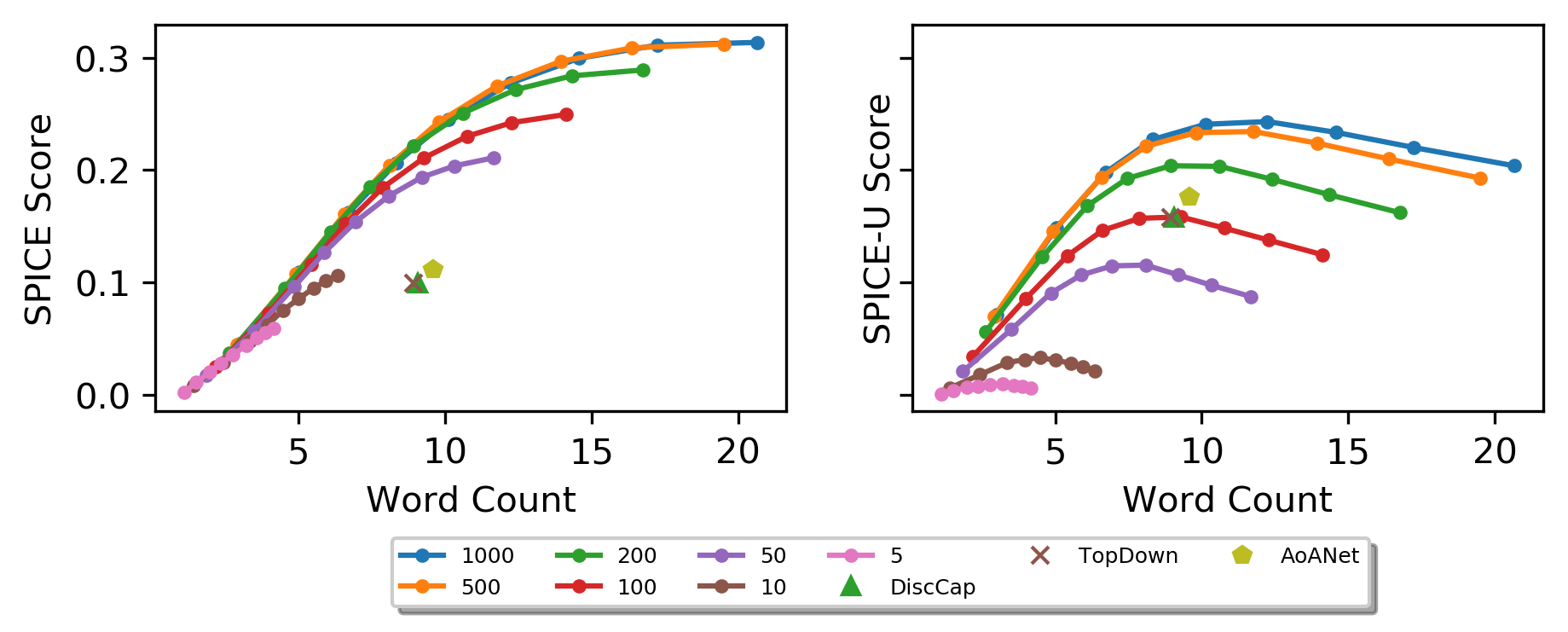}
    \caption{Comparison of state of the art TopDown model~\cite{anderson2018bottom}, DiscCap model~\cite{luo2018discriminability}, AoANet model~\cite{huang_attention_2019}, and our object detection-based models (best viewed in color). The $x$-axis is the average caption length in words. The $y$-axis is the SPICE score~\cite{spice2016} (left) and proposed SPICE-U score (right) on 1,076 images (the intersection of the COCO~\cite{lin2014coco} and the Visual Genome dataset~\cite{krishna2017visual} which not appear in the training set of both object detection and captioning models). The different curves of the object-based model correspond to running different numbers of object detectors (e.g., detecting only the 1000, 500, etc most common object classes in the image) and producing simplistic captions of the form ``There is a tennis ball, court and person''. For each curve, performance is shown across varying detection thresholds from 0.1 to 0.9. A simple object-based model that only outputs the 10 most common object classes seen in images (brown) outperforms a state of the art discriminative captioning model (green triangle) on SPICE, but not on SPICE-U.}
    \label{fig:fewobjects}
\end{figure}

\smallsec{Findings}
We run a simple experiment to show that the SPICE metric can be fooled by very simple baseline models that only recognize the {\bf  10} \emph{most common object classes} in images, and nothing else!\footnote{The objects classes are: man, person, tree, ground, shirt, wall, sky, window, building, and head.} To do so, we design an object-based captioning model consisting of a set of object detectors. The object detectors are trained jointly as a Faster R-CNN model~\cite{ren2015faster}, on the Visual Genome training dataset~\cite{krishna2017visual}.\footnote{The trained object detectors are taken from the bottom-up part of the captioning model~\cite{anderson2018bottom}.} Given a set of detected objects such as ``tennis ball,'' ``court'' and ``person,'' the final caption is generated following a template as: ``There is a tennis ball, court and person''.\footnote{The resulting model is similar to \textit{Baby Talk}~\cite{kulkarni2011baby}, which uses object, attribute, and relationship classifiers to generate image descriptions.} We evaluate the accuracy of this system using the SPICE (Eqn.~\ref{eq:spice}). The evaluation is done on 1,076 images (the intersection of the COCO~\cite{lin2014coco} and the Visual Genome dataset~\cite{krishna2017visual} which not appear in the training set of both object detection and captioning models) using their ground truth concept annotations from Visual Genome.

To help interpret the results, we compare this baseline model with three modern captioning systems: the bottom-up and top-down attention model~\cite{anderson2018bottom}, which combines the bottom-up region features generated from object detector with top-down attention mechanism, the model of Luo et al.~\cite{luo2018discriminability}, which includes a ``discriminability'' loss to encourage unique captions, and the model of Huang et al.~\cite{huang_attention_2019}, which extends conventional models with a stronger attention mechanism. The models are trained on the COCO dataset~\cite{lin2014coco} with the split of~\cite{karpathy2015deep}. Figure~\ref{fig:fewobjects} (left) details the results of the experiment. Surprisingly, according to this metric an object detector that only knows {\bf 10} object classes rivals a state of the art captioning model: our object-based captioning model achieves a SPICE score of 0.11 versus 0.10 of \cite{luo2018discriminability}! This occurs even despite producing fewer words on average per caption: 6.4 versus 9.1. Further, we observe that given access to a (still limited) set of 500 object detectors, our simple baseline produces significantly higher SPICE scores ($\geq 0.3$).

\smallsec{Conclusions} These surprising findings are likely due to two reasons. First, the SPICE score  gives equal weight to different concepts.  This means that, for example, a caption that names generic objects like ``tree'' and ``person'' scores the same as a caption that identifies the two unique objects in the image, such as ``volleyball'' or ``gazebo'', giving a perhaps unfair advantage to our simple baseline. We will address this by proposing a new uniqueness-based metric in Section~\ref{sec:uniqueness-metric}. Second, modern captioning systems are optimized to rely too heavily on the common concepts, failing to fully leverage their image understanding capabilities, and we propose some strategies to mitigate that in Section~\ref{sec:methods}.


\section{SPICE-U: A uniqueness-aware metric}
\label{sec:uniqueness-metric}

Inspired by the observations in Section~\ref{sec:analysis}, we introduce the SPICE-U metric (``Semantic Propositional Image Caption Evaluation with Uniqueness'') to encourage captions to capture the diversity and uniqueness of real-world images. 

\smallsec{Uniqueness}
We define the uniqueness of a single visual concept $p$ as:
\begin{equation}
    \mathrm{Un}(p) = \frac{\mathrm{\# \; images \; not \; containing \; p}}
    {\mathrm{\# \; images \; total}}
\end{equation}
This is similar to the notion of \textit{inverse document frequency} (IDF) in text retrieval~\cite{sparck1972statistical}, which allows for weighting down common words in text. While this concept is also used in CIDEr~\cite{vedantam2015cider}, they compute IDF over n-grams, not visual concepts. Note that our definition of uniqueness is complementary to saliency -- while saliency measures how prominent a concept is in the image, uniqueness aims to identify parts of the image that make it \emph{interesting}. Future work could involve investigating combinations of these.

For computational tractability, we approximate the denominator using a large set of images (e.g. the training set). For example, if $p$ is $tree$, contained in 28,186 of 113,287 images in the COCO training set~\cite{lin2014coco}, $\mathrm{Un}[tree] = 0.75$. We realize that this approximation introduces some dependence on the corpus, but this is similar to calculating IDF using a large text corpus in metrics like CIDEr. Further, even measures like recall implicitly make corpus-specific assumptions, e.g. by considering the set of ground truth concepts to be those concepts seen in the dataset. 

To define the uniqueness of a set of predictions $\pred$, we want to consider the uniqueness of its constituent concepts. One natural definition would be:
\begin{equation}
    \mathrm{Un}(\pred)=\sum_{p\in\pred}\mathrm{Un}(p)
\end{equation}
However, this definition is undesirable for several reasons. First, it's not between 0 and 1, making it difficult to reason about in comparison with precision and recall. Second, and more problematically, it increases with every additional concept (unless the concept is present in 100\% of the training images), encouraging long captions. Finally, it encourages the models to make incorrect predictions and detect unusual concepts not present in the image just to increase the uniqueness score.

Instead, we use a definition that measures the uniqueness of a set of predictions compared to the best (most unique) set of predictions which could have been made. To do so, consider alternative predictions $\mathbf{A}$ of the same length as $\pred$. As to not encourage a reduction in accuracy through uniqueness, we further assume $\mathbf{A}$ consists only of the concepts that appear either within $\pred$ or within the ground truth set $\gt$. Concretely:
\begin{equation}
\mathcal{A}({\pred;\gt}) = \{\mathbf{A}: \mathbf{A} \in \gt \cup \pred, |\mathbf{A}|=|\pred|\}
\end{equation}
For example, if the image contains a $cat$ and a $dog$, and the prediction was $cat$ and $fish$:
\begin{equation}
    \mathcal{A}(\{(cat, fish)\}, \{(cat,dog)\}) = \{(cat,dog),(cat,fish),(dog,fish)\}
\end{equation}
Given this definition, we then define the \emph{uniqueness} of a prediction as:
\begin{align}
\label{eq:uniqueness}
    \mathrm{Uniq}(\pred;\gt)=
\frac{
          \mathrm{Un}(\pred)
       - \min_{\mathbf{A} \in \mathcal{A}(\gt;\pred)}
            \mathrm{Un}(\mathbf{A}) 
}{
  \max_{\mathbf{A} \in \mathcal{A}(\gt;\pred)}
        \mathrm{Un}(\mathbf{A})
    - \min_{\mathbf{A} \in \mathcal{A}(\gt;\pred)}
        \mathrm{Un}(\mathbf{A})
        } 
\end{align}
Intuitively, this measures how unique the caption is compared to others of the same length that could have been conceivably generated. For example, consider an image that contains a $person$ (uniqueness score of 0.75), $table$ (score of 0.87), and $elephant$ (score of 0.98). If the model captions only one of these objects and nothing else, it will be rewarded with a uniqueness score of $1$ if the object it chooses is $elephant$, $0$ if it outputs $person$, and $0.52$ if it outputs $table$. Note that predicting a more unique, yet incorrect, object would not give the model an additional reward. Similarly, if the image did not contain an $elephant$, then the model would receive the full uniqueness score of $1$ for predicting the most unique object $table$. This ensures that models are rewarded for noticing unique things in the image but not unfairly penalized on images with only common concepts.

\smallsec{Combined metric} The uniqueness-aware measure of the quality of a caption is then a combination through harmonic mean of SPICE (Eqn.~\ref{eq:spice}), and uniqueness (Eqn.~\ref{eq:uniqueness}): 

\begin{align}
\label{eq:SPICE-U}
\mathrm{SPICE\mymathhyphen U}(\pred;\gt)=   \frac{2}{1/\mathrm{SPICE}(\pred;\gt) + 1/\mathrm{Uniq}(\pred;\gt)} 
\end{align}

Consider the example above of an image that contains a $person$, $table$ and $elephant$, and two captions: ``There is a table'' and ``There is an elephant.'' The original SPICE score of Eqn.~\ref{eq:spice} would be $0.5$ for both captions (recall $1/3$, precision $1$), failing to recognize that one is a much more useful caption than the other. However, SPICE-U score would be $0.67$ for ``There is an elephant'' and $0.51$ for ``There is a table,'' correctly selecting the most informative description.\footnote{For ``There is a person'' uniqueness is $0$, since it's the most common of the objects, and SPICE-U score is 0 by definition.}


\begin{table}[t]
	\begin{center}
		\caption{Evaluation of various metrics against human judgements. First five columns show pairwise judgment accuracy with fifty reference captions on the PASCAL-50 dataset (HC: both sentences written by humans for the corresponding image,  HI: both sentences written by humans -- one for the corresponding image and one for a random image, HM: one caption written by human and another generated by a model, MM: captions generated by two different models.) The last column is Pearson's correlation between human preferences and each metric on images from PASCAL-50.}
		\label{tab:human}
		
		\begin{tabular}{lccccc|c}
			\toprule
			& HC & HI & HM & MM & ALL & Pearson’s\\
			\midrule 
			BLEU-4 & 55.00 & 97.30 & 92.60 & 61.80 & 76.68 & 0.581\\
			ROUGE & 54.60 & 98.70 & 96.00 & 62.00 & 77.83 & 0.732\\
			METEOR & 57.50 & \bf{99.30} & \bf{96.90} & 62.30 & 79.00 & 0.710\\
			CIDEr & 53.00 & \bf{99.30} & 92.10 & 67.10 & 77.88 & 0.641\\
			SPICE  & \bf{66.80} & 98.50 & 93.80 & \bf{71.10} & 82.55 & 0.749\\
			SPICE-U & 66.50 & 98.60 & 94.40 & 70.80 & \bf{82.58} & \bf{0.767}\\
			\bottomrule
		\end{tabular}
	\end{center}
\end{table}

\smallsec{Advantage of SPICE-U}
We follow the setup of \cite{spice2016} to analyze correlation of SPICE-U with human judgements when determining the similarity of sentences. We use the PASCAL-50S dataset~\cite{vedantam2015cider}, which contains 50 ground truth captions for each image. Human annotators were provided with a pair of candidate sentences ($b$, $c$) and asked which was more similar to sentence $a$, which is one of the ground truth captions for an image. Consider an image with a set of ground truth captions $A=\{a_k\}$ and a reference pair of sentences $(b,c)$ as above, where without loss of generality we assume that humans favored $b$ over $c$ for this image (i.e., on average over all $a_k$, humans found $a_k$ to be more similar to $b$ than $c$). We say that a $\mathrm{metric}$ agrees with humans if $\mathrm{metric}(b,A) \geq \mathrm{metric}(c,A)$. From table~\ref{tab:human}, we observe that SPICE-U achieves better judgement accuracy than other metrics and comparable accuracy with SPICE, especially outperforming SPICE on HM pairs. This shows that SPICE-U can indeed capture the diverse nature of human written captions and can help separate two captions that are both correct but differ in quality. Despite being a standard test on PASCAL-50S, measuring the accuracy abstracts away detailed human preferences, and causes issue when two candidate captions get similar human votes. To mitigate this, we also evaluate Pearson's correlation between human preferences and each metric\footnote{We calculate the correlation between the mean value of human votes (+1 if they prefer caption b over caption c, -1 otherwise) and the score $R_m(b) - R_m(c)$, where $R_m(s)$ is the score of sentence $s$ given by metric $m$.}. SPICE-U achieves the best correlation score among all metrics.


\section{Generating unique and informative captions}
\label{sec:methods}

SPICE-U aims to capture the uniqueness of a particular caption given an image. Intuitively, any captioning model that maximizes SPICE-U must forge a strong connection between the semantic concepts in the image and the linguistic concepts in the caption it generates. However, in the predominant (current) regime of end-to-end training with loss functions such as cross entropy, there is no explicit objective which enables this connection.

Formally, current captioning models decode using the following objective:
\begin{equation}
\hat{s} = \arg \max_s \log P(s | I; \theta)
\end{equation}
where $s$ is the caption, $I$ is the image and $\theta$ are the learned parameters of the model.
However, this ignores the dependency from the caption to the image $P(I|s)$, which is critical for ensuring that the caption adequately (and uniquely) describes the image. A similar observation was made in machine translation~\cite{li2016diversity,li2016mutual} where the input and output are sentences in two different languages.

One solution to this problem is to maximize mutual information (MMI) instead of cross-entropy:
\begin{align}
\begin{split}
\hat{s} &= \arg \max_s \log \frac{P(I, s)}{P(I)P(s)^\lambda} \\
&= \arg \max_s  \log P(s | I) - \lambda \log P(s) \\
&= \arg \max_s (1-\lambda)\log P(s|I) +\lambda \log P(I|s)
\end{split}
\label{eq:MMI}
\end{align}

\begin{algorithm}[t]
	\caption{Generating caption with beam decoding and re-ranking}
	\begin{algorithmic}[1]
		\REQUIRE Caption model with parameter $\theta_c$, language model with parameter $\theta_l$, image $I$, weighting factor $\lambda$
		
		\ENSURE Generated caption $s$
		
		\STATE Beam decode top-k captions $\{s^{(1)}, ..., s^{(k)}\}$ along with probabilities $\{P(s^{(1)}|I;\theta_c), ..., P(s^{(k)}|I;\theta_c)\}$ with caption model 
		
		\STATE Generate probabilities for entire captions $\{P(s^{(1)};\theta_l), ..., P(s^{(k)};\theta_l)\}$ with language model
		
		\STATE $s \leftarrow \arg\max_{s^{(i)}} ~\log P(s^{(i)}|I;\theta_c) - \lambda \log P(s^{(i)};\theta_l)$
	\end{algorithmic}
	\label{alg:rerank}
\end{algorithm}

However, since training a model to predict $P(I|s)$ is not trivial~\cite{li_object-driven_2019,zhang2017stackgan,AttnGAN}, we propose to use second line in the MMI objective above to \textit{re-rank captions} produced by a standard beam decoding mechanism. To this end, we train language models to obtain likelihood estimates for captions, $\log P(s) = \sum_i \log P(s_i | s_{<i})$. In particular, we investigate three variants of language models:
\begin{enumerate}
	\item \textbf{Unigram LM:} A simple unigram language model estimated from the train set, $P(s) = \prod_{i} P(s_i)$
	\item \textbf{LSTM LM:} An LSTM language model trained on captions in the train set. 
	\item \textbf{Interpolated LM:} A log-linear interpolation\footnote{We also tried linear interpolation and it works not as good as the log-linear interpolation.} between the variants above: \\
	\begin{equation}
	\label{eq:interpolation}
		P_{int}(s_i | s_{<i}) = P_{uni}(s_i)^\alpha P_{LSTM}(s_i | s_{<i})^{1-\alpha}
	\end{equation}

\end{enumerate}
We generate the top-$k$ captions using the baseline model and then re-rank them using their newly computed scores, described in Algorithm~\ref{alg:rerank}.


\section{Experiments}

\smallsec{Data} 
We conduct experiments on the COCO \cite{lin2014coco} dataset which contains images of everyday scenes with common objects in their natural context. For captioning task, every image is annotated with five human  captions, mostly short sentences summarizing the important parts of the scene. We adopt the popular split of this dataset from Karpathy et al.~\cite{karpathy2015deep}, which contains 113,287 images for training and 5,000 images for validation and test respectively.

\smallsec{Model}
We use three recent captioning models as our baselines. The bottom-up and top-down attention model (TopDown) from Anderson et al.~\cite{anderson2018bottom} utilizes object detector to propose salient image regions as bottom-up features and then uses top-down attention to decide weight for each region. The discriminative captioning model (DiscCap) from Luo et al.~\cite{luo2018discriminability} is trained explicitly with proposed `discriminability' loss besides standard cross-entropy loss to encourage unique captions that can distinguish between different images. The attention on attention model (AoANet) from Huang et al.~\cite{huang_attention_2019} extends conventional attention mechanism with another attention to determine the relevance between attention results and queries. We use off-the-shelf implementations for these models\footnote{The TopDown model from \url{https://github.com/poojahira/image-captioning-bottom-up-top-down}, the DiscCap from \url{https://github.com/ruotianluo/DiscCaptioning} and AoANet from \url{https://github.com/husthuaan/AoANet}.}. For language model, we train a one-layer LSTM with hidden size of 512 and embedding size of 300.

\smallsec{Re-ranking} 
We use the captioning model with beam decoding to generate top 10 candidates along with probabilities $P(s|I)$ for re-ranking. The language model is then used to generate the $P(s)$ for each candidate caption and finally the caption with the maximum mutual information is selected according to Eqn.~\ref{eq:MMI} as the predicted caption. 

The hyperparameters $\lambda$ (language model weight in Eqn.~\ref{eq:MMI}) and $\alpha$ (coefficient in interpolation model, Eqn.~\ref{eq:interpolation}) are selected for each model on the validation set using a grid search (0 to 1, step size of 0.1). 

We cross-validate with the objective of optimizing the geometric mean\footnote{The captioning metrics measure different aspects of the captions and are largely uncorrelated with each other~\cite{rohrbach2018object}; we use the geometric mean as a simple summary statistic of the overall performance of the models. For CHAIR lower scores are better so we use $\frac{1}{CHAIR}$ in the geometric mean.} across several evaluation metrics (BLEU-4, METEOR, CIDEr, CHAIRs, SPICE and SPICE-U). The resulting hyperparameters are: $\lambda=0.3$ on TopDown+Unigram, $\lambda=0.2$ on TopDown+LSTM, $\lambda=0.4,\alpha=0.8$ on TopDown+Interpolated, $\lambda=1.0$ on DiscCap+Unigram, $\lambda=0.1$ on DiscCap+LSTM, $\lambda=0.8,\alpha=0.9$ on DiscCap+Interpolated, and $\lambda=0.4$ on AoANet+Unigram, $\lambda=0.1$ on AoANet+LSTM, $\lambda=0.5,\alpha=0.9$ on AoANet+Interpolated.

\begin{table*}[t]
	\begin{center}
		\scriptsize
		\caption{Comparison of three different state-of-the-art captioning systems~\cite{anderson2018bottom,luo2018discriminability,huang_attention_2019}, along with our proposed re-ranking schemes, evaluated using different metrics on the COCO test split from~\cite{karpathy2015deep}.}
		\label{tab:test performance}
		\begin{tabular}{lcccccc|c}
			& BLEU & METEOR & CIDEr & CHAIRs ($\downarrow$)  & SPICE & SPICE-U & GeoMean
			\\
			\midrule
			
			TopDown~\cite{anderson2018bottom} &  \bf{23.03} & 28.98 & \bf{108.13} & 8.68 & 20.62 & 23.70 & 12.63\\
			TopDown+Unigram & 22.88 & \bf{29.06} & 107.04 & 8.10 & \bf{20.82} & 25.05 & 12.89\\
			TopDown+LSTM & 22.79 & 28.48 & 107.59 & 8.20 & 20.52 & 24.46 & 12.74\\
			TopDown+Interpolated & 22.77 & 28.84 & 106.42 & \bf{7.80} & 20.72 & \bf{25.27} & \bf{12.94} \\
			
			\midrule
			DiscCap~\cite{luo2018discriminability} & \bf{21.93} & \bf{27.55} & \bf{112.39} & 11.92 & \bf{20.32} & 23.74 & 11.84\\
			DiscCap+Unigram & 21.56 & 27.38 & 110.41 & 10.88 & 20.28 & 24.60 & 12.00\\
			DiscCap+LSTM & 21.64 & 27.40 & 111.73 & 11.34 & 20.17 & 23.79 & 11.87\\
			DiscCap+Interpolated & 21.58 & 27.42 & 110.90 & \bf{10.84} & 20.27 & \bf{24.52} & \bf{12.02}\\
			
			\midrule
			AoANet~\cite{huang_attention_2019} & \bf{27.53} & 30.37 & \bf{129.12} & 10.40 & 22.77 & 26.04 & 13.54\\
			AoANet+Unigram & 27.30 & \bf{30.43} & 128.66 & 9.52 & 22.79 & 26.46 & 13.75\\
			AoANet+LSTM & 27.36 & 30.26 & 128.79 & 10.24 & 22.71 & 26.12 & 13.55\\
			AoANet+Interpolated & 27.18 & 30.39 & 128.15 & \bf{9.28} & \bf{22.81} & \bf{26.53} & \bf{13.80}\\
			
			\bottomrule
			
		\end{tabular}
	\end{center}
\end{table*}

\smallsec{Results}
Table~\ref{tab:test performance} summarizes the results. For the TopDown baseline, the TopDown+Interpolated modification improves SPICE-U by an absolute $1.6\%$ over the baseline (from $23.7\%$ to $25.3\%$) and the geometric mean over all metrics by a relative $2.4\%$ (from $12.6\%$ to $12.9\%$). 
For DiscCap model, DiscCap+Interpolated led to an absolute improvement of $0.8\%$ on SPICE-U (from $23.7\%$ to $24.5\%$) and $1.7\%$ relative on the geometric mean (from $11.8\%$ to $12.0\%$).
For AoANet, AoANet+Interpolated improves SPICE-U by an absolute $0.5\%$ (from $26.0\%$ to $26.5\%$) and a relative $2.2\%$ improvement on geometric mean.

\begin{figure*}[t]
	\centering
	\input{img/qual_figure_new}
	\caption{Captions generated by the AoANet~\cite{huang_attention_2019} model (in italics) and by our variation AoANet+Interpolated (in regular font). The modification we introduce encourages the model to output more descriptive and accurate captions, such as describing the place (``beach''), the type of the sign (``no parking sign''), the presence of a prominent object (``paper container'', ``statue'') in the first four images. However, there are also some images (like the last one) where despite improvements in SPICE-U the changes are less interesting, such as simply replacing ``group'' with ``herd''.} 
	\label{fig:qual_results}
\end{figure*}
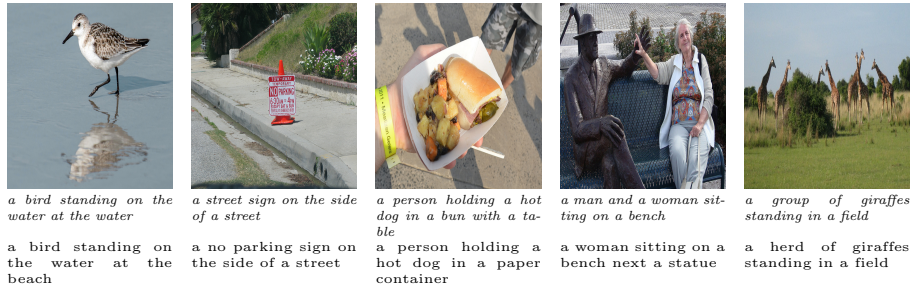

Figure~\ref{fig:qual_results} shows qualitative examples: as expected, the updated captions correspond to more detailed descriptions of the image. The improvements demonstrated here are the result of quite simple algorithmic modification yet propose a promising path forward for improving modern image captioning system.

\section{Conclusion}
\label{sec:conclusion}

State of the art image captioning models produce generic captions, leaving out important image details and misrepresenting facts. In this paper, we quantitatively demonstrated that both modern captioning systems and evaluation metrics tend towards generating and rewarding captions with commonly occurring concepts from the training data. We then introduced a new notion of \textit{uniqueness} and used it to propose a new metric, SPICE-U. Our studies show that SPICE-U correlates better with human judgements compared to SPICE. Finally, we utilized the notion of maximizing mutual information to re-rank captions produced by any captioning system. Our experiments demonstrate that our method results in unique and informative captions, and yields promising improvements over three different state-of-the-art models.

\subsubsection{Acknowledgments.} This work is partially supported by KAUST under Award No. OSRCRG2017-3405, by Samsung and by the Princeton CSML DataX award. 
We would like to thank Arjun Mani, Vikram Ramaswamy and Angelina Wang for their helpful feedback on the paper.
 
\bibliographystyle{splncs04}
\bibliography{egbib}
\end{document}

%% file: img/qual_figure_new.tex
{\tiny
\begin{tabular}{p{0.18\linewidth}@{\hskip0.1in}p{0.18\linewidth}@{\hskip0.1in}p{0.18\linewidth}@{\hskip0.1in}p{0.18\linewidth}@{\hskip0.1in}p{0.18\linewidth}}

\includegraphics[width=\linewidth,height=70pt]{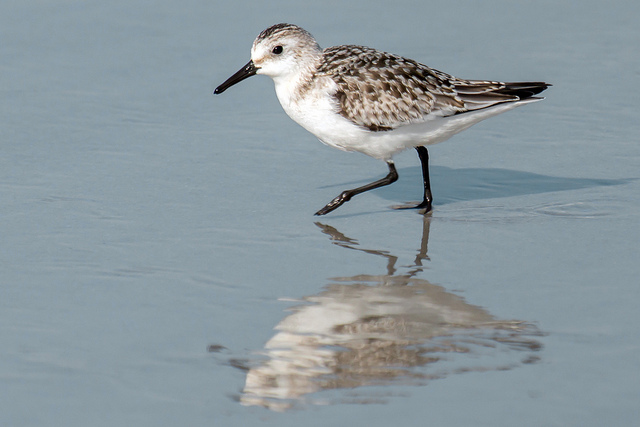}
&
\includegraphics[width=\linewidth,height=70pt]{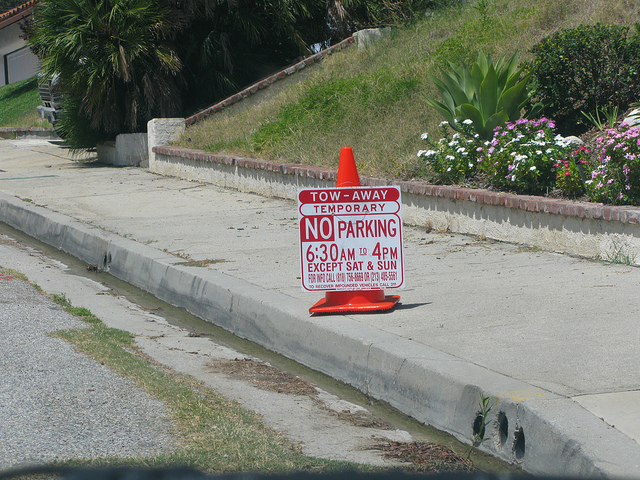}
&
\includegraphics[width=\linewidth,height=70pt]{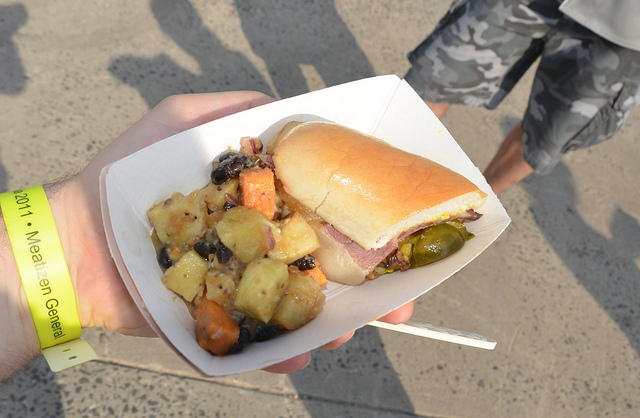}
&
\includegraphics[width=\linewidth,height=70pt]{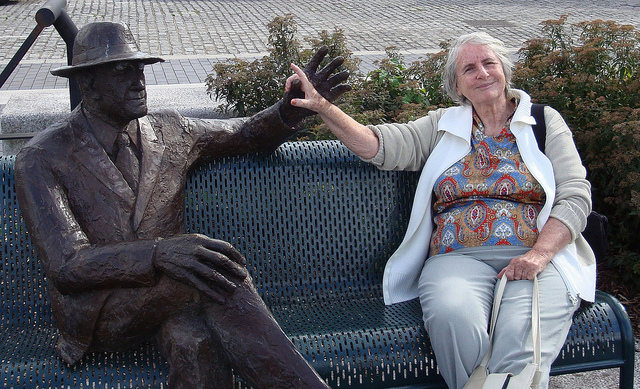}
&
\includegraphics[width=\linewidth,height=70pt]{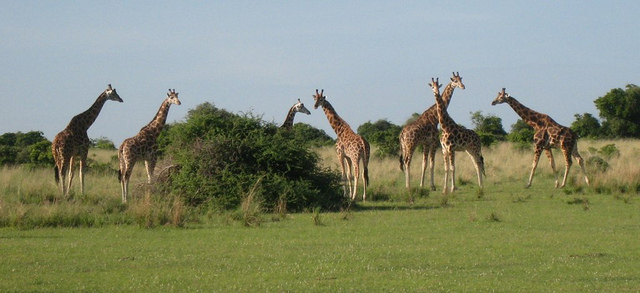}
\\
\textit{a bird standing on the water at the water}
&
\textit{a street sign on the side of a street}
&
\textit{a person holding a hot dog in a bun with a table}
&
\textit{a man and a woman sitting on a bench}
&
\textit{a group of giraffes standing in a field}
\\
a bird standing on the water at the beach
&
a no parking sign on the side of a street
&
a person holding a hot dog in a paper container
&
a woman sitting on a bench next a statue
&
a herd of giraffes standing in a field
\\

\end{tabular}
}

%% file: main.bbl
\begin{thebibliography}{10}
\providecommand{\url}[1]{\texttt{#1}}
\providecommand{\urlprefix}{URL }
\providecommand{\doi}[1]{https://doi.org/#1}

\bibitem{spice2016}
Anderson, P., Fernando, B., Johnson, M., Gould, S.: {SPICE}: {Semantic}
  {Propositional} {Image} {Caption} {Evaluation}. In: {European} {Conference}
  on {Computer} {Vision} (ECCV) (2016)

\bibitem{anderson2018bottom}
Anderson, P., He, X., Buehler, C., Teney, D., Johnson, M., Gould, S., Zhang,
  L.: Bottom-{Up} and {Top}-{Down} {Attention} for {Image} {Captioning} and
  {Visual} {Question} {Answering}. In: 2018 {IEEE}/{CVF} {Conference} on
  {Computer} {Vision} and {Pattern} {Recognition} (CVPR) (2018)

\bibitem{bahl1986maximum}
Bahl, L., Brown, P., de~Souza, P., Mercer, R.: Maximum mutual information
  estimation of hidden {Markov} model parameters for speech recognition. In:
  {IEEE} {International} {Conference} on {Acoustics}, {Speech}, and {Signal}
  {Processing} (ICASSP) (1986)

\bibitem{cui2018learning}
Cui, Y., Yang, G., Veit, A., Huang, X., Belongie, S.: Learning to {Evaluate}
  {Image} {Captioning}. In: 2018 {IEEE}/{CVF} {Conference} on {Computer}
  {Vision} and {Pattern} {Recognition} (CVPR) (2018)

\bibitem{datta2017multimodal}
Datta, D., Varma, S., Chowdary~C., R., Singh, S.K.: Multimodal {Retrieval}
  using {Mutual} {Information} based {Textual} {Query} {Reformulation}. Expert
  Systems with Applications  (2017)

\bibitem{dognin2019adversarial}
Dognin, P., Melnyk, I., Mroueh, Y., Ross, J., Sercu, T.: Adversarial {Semantic}
  {Alignment} for {Improved} {Image} {Captions}. In: 2019 {IEEE}/{CVF}
  {Conference} on {Computer} {Vision} and {Pattern} {Recognition} ({CVPR})
  (2019)

\bibitem{henning2018estimating}
Henning, C.A., Ewerth, R.: Estimating the {Information} {Gap} between {Textual}
  and {Visual} {Representations}. In: Proceedings of the 2017 {ACM} on
  {International} {Conference} on {Multimedia} {Retrieval} (2017)

\bibitem{huang_attention_2019}
Huang, L., Wang, W., Chen, J., Wei, X.Y.: Attention on {Attention} for {Image}
  {Captioning}. In: 2019 {IEEE}/{CVF} {International} {Conference} on
  {Computer} {Vision} ({ICCV}). IEEE (2019)

\bibitem{johnson2016densecap}
Johnson, J., Karpathy, A., Fei-Fei, L.: {DenseCap}: {Fully} {Convolutional}
  {Localization} {Networks} for {Dense} {Captioning}. In: 2016 {IEEE}
  {Conference} on {Computer} {Vision} and {Pattern} {Recognition} ({CVPR})
  (2016)

\bibitem{johnson2015image}
Johnson, J., Krishna, R., Stark, M., Li, L.J., Shamma, D.A., Bernstein, M.S.,
  Fei-Fei, L.: Image retrieval using scene graphs. In: 2015 {IEEE} {Conference}
  on {Computer} {Vision} and {Pattern} {Recognition} ({CVPR}) (2015)

\bibitem{karpathy2015deep}
Karpathy, A., Fei-Fei, L.: Deep visual-semantic alignments for generating image
  descriptions. In: 2015 {IEEE} {Conference} on {Computer} {Vision} and
  {Pattern} {Recognition} ({CVPR}) (2015)

\bibitem{kimura2019selecting}
Kimura, R., Iida, S., Cui, H., Hung, P.H., Utsuro, T., Nagata, M.: Selecting
  {Informative} {Context} {Sentence} by {Forced} {Back}-{Translation}. In:
  Proceedings of {Machine} {Translation} {Summit} {XVII} {Volume} 1: {Research}
  {Track} (2019)

\bibitem{krause2017hierarchical}
Krause, J., Johnson, J., Krishna, R., Fei-Fei, L.: A {Hierarchical} {Approach}
  for {Generating} {Descriptive} {Image} {Paragraphs}. In: 2017 {IEEE}
  {Conference} on {Computer} {Vision} and {Pattern} {Recognition} ({CVPR})
  (2017)

\bibitem{krishna2017visual}
Krishna, R., Zhu, Y., Groth, O., Johnson, J., Hata, K., Kravitz, J., Chen, S.,
  Kalantidis, Y., Li, L.J., Shamma, D.A., Bernstein, M.S., Fei-Fei, L.: Visual
  {Genome}: {Connecting} {Language} and {Vision} {Using} {Crowdsourced} {Dense}
  {Image} {Annotations}. International Journal of Computer Vision (1) (2017)

\bibitem{kulkarni2011baby}
Kulkarni, G., Premraj, V., Ordonez, V., Dhar, S., Li, S., Choi, Y., Berg, A.C.,
  Berg, T.L.: {BabyTalk}: {Understanding} and {Generating} {Simple} {Image}
  {Descriptions}. IEEE Transactions on Pattern Analysis and Machine
  Intelligence (12) (2013)

\bibitem{banerjee2005meteor}
Lavie, A., Agarwal, A.: Meteor: an automatic metric for {MT} evaluation with
  high levels of correlation with human judgments. In: Proceedings of the
  {Second} {Workshop} on {Statistical} {Machine} {Translation} (2007)

\bibitem{li2016diversity}
Li, J., Galley, M., Brockett, C., Gao, J., Dolan, B.: A {Diversity}-{Promoting}
  {Objective} {Function} for {Neural} {Conversation} {Models}. In: Proceedings
  of the 2016 {Conference} of the {North} {American} {Chapter} of the
  {Association} for {Computational} {Linguistics}: {Human} {Language}
  {Technologies} (2016)

\bibitem{li2016mutual}
Li, J., Jurafsky, D.: Mutual {Information} and {Diverse} {Decoding} {Improve}
  {Neural} {Machine} {Translation}. arXiv:1601.00372 [cs]  (2016), arXiv:
  1601.00372

\bibitem{li_object-driven_2019}
Li, W., Zhang, P., Zhang, L., Huang, Q., He, X., Lyu, S., Gao, J.:
  Object-{Driven} {Text}-{To}-{Image} {Synthesis} via {Adversarial} {Training}.
  In: 2019 {IEEE}/{CVF} {Conference} on {Computer} {Vision} and {Pattern}
  {Recognition} ({CVPR}) (2019)

\bibitem{lin2014coco}
Lin, T.Y., Maire, M., Belongie, S., Hays, J., Perona, P., Ramanan, D., Dollár,
  P., Zitnick, C.L.: Microsoft {COCO}: {Common} {Objects} in {Context}. In:
  {European} {Conference} on {Computer} {Vision} (ECCV) (2014)

\bibitem{lindh2018generating}
Lindh, A., Ross, R.J., Mahalunkar, A., Salton, G., Kelleher, J.D.: Generating
  {Diverse} and {Meaningful} {Captions}. In: Artificial {Neural} {Networks} and
  {Machine} {Learning} ({ICANN}) (2018)

\bibitem{liu2019generating}
Liu, L., Tang, J., Wan, X., Guo, Z.: Generating {Diverse} and {Descriptive}
  {Image} {Captions} {Using} {Visual} {Paraphrases}. In: 2019 {IEEE}/{CVF}
  {International} {Conference} on {Computer} {Vision} ({ICCV}) (2019)

\bibitem{liu2017improved}
Liu, S., Zhu, Z., Ye, N., Guadarrama, S., Murphy, K.: Improved {Image}
  {Captioning} via {Policy} {Gradient} optimization of {SPIDEr}. In: 2017
  {IEEE} {International} {Conference} on {Computer} {Vision} ({ICCV}) (2017)

\bibitem{liu2018show}
Liu, X., Li, H., Shao, J., Chen, D., Wang, X.: Show, {Tell} and {Discriminate}:
  {Image} {Captioning} by {Self}-retrieval with {Partially} {Labeled} {Data}.
  In: {European} {Conference} on {Computer} {Vision} (ECCV) (2018)

\bibitem{lu2018entity}
Lu, D., Whitehead, S., Huang, L., Ji, H., Chang, S.F.: Entity-aware {Image}
  {Caption} {Generation}. In: Proceedings of the 2018 {Conference} on
  {Empirical} {Methods} in {Natural} {Language} {Processing} ({EMNLP}) (2018)

\bibitem{lu_knowing_2017}
Lu, J., Xiong, C., Parikh, D., Socher, R.: Knowing {When} to {Look}: {Adaptive}
  {Attention} via a {Visual} {Sentinel} for {Image} {Captioning}. In: 2017
  {IEEE} {Conference} on {Computer} {Vision} and {Pattern} {Recognition}
  ({CVPR}) (2017)

\bibitem{luo2018discriminability}
Luo, R., Shakhnarovich, G., Cohen, S., Price, B.: Discriminability {Objective}
  for {Training} {Descriptive} {Captions}. In: 2018 {IEEE}/{CVF} {Conference}
  on {Computer} {Vision} and {Pattern} {Recognition} (CVPR) (2018)

\bibitem{Mao_2016_CVPR}
Mao, J., Huang, J., Toshev, A., Camburu, O., Yuille, A., Murphy, K.: Generation
  and {Comprehension} of {Unambiguous} {Object} {Descriptions}. In: 2016 {IEEE}
  {Conference} on {Computer} {Vision} and {Pattern} {Recognition} ({CVPR})
  (2016)

\bibitem{melas2018training}
Melas-Kyriazi, L., Rush, A., Han, G.: Training for {Diversity} in {Image}
  {Paragraph} {Captioning}. In: Proceedings of the 2018 {Conference} on
  {Empirical} {Methods} in {Natural} {Language} {Processing} ({EMNLP}) (2018)

\bibitem{papineni2002bleu}
Papineni, K., Roukos, S., Ward, T., Zhu, W.J.: {BLEU}: a method for automatic
  evaluation of machine translation. In: Proceedings of the 40th {Annual}
  {Meeting} on {Association} for {Computational} {Linguistics} (ACL) (2001)

\bibitem{povey2002minimum}
Povey, D., Woodland, P.: Minimum {Phone} {Error} and {I}-smoothing for improved
  discriminative training. In: 2002 {IEEE} {International} {Conference} on
  {Acoustics}, {Speech}, and {Signal} {Processing} (2002), iSSN: 1520-6149

\bibitem{ren2015faster}
Ren, S., He, K., Girshick, R., Sun, J.: Faster {R}-{CNN}: {Towards}
  {Real}-{Time} {Object} {Detection} with {Region} {Proposal} {Networks}. IEEE
  Transactions on Pattern Analysis and Machine Intelligence (6) (2017)

\bibitem{rohrbach2018object}
Rohrbach, A., Hendricks, L.A., Burns, K., Darrell, T., Saenko, K.: Object
  {Hallucination} in {Image} {Captioning}. In: Proceedings of the 2018
  {Conference} on {Empirical} {Methods} in {Natural} {Language} {Processing}
  (EMNLP) (2018)

\bibitem{shetty2017speaking}
Shetty, R., Rohrbach, M., Hendricks, L.A., Fritz, M., Schiele, B.: Speaking the
  {Same} {Language}: {Matching} {Machine} to {Human} {Captions} by
  {Adversarial} {Training}. In: 2017 {IEEE} {International} {Conference} on
  {Computer} {Vision} ({ICCV}) (2017)

\bibitem{sparck1972statistical}
Sparck~Jones, K.: A {Statistical} {Interpretation} of {Term} {Specificity} and
  {Its} {Application} in {Retrieval}. Journal of Documentation (1) (1972)

\bibitem{tu2017neural}
Tu, Z., Liu, Y., Shang, L., Liu, X., Li, H.: Neural {Machine} {Translation}
  with {Reconstruction}. In: Thirty-{First} {AAAI} {Conference} on {Artificial}
  {Intelligence} (2017)

\bibitem{vedantam2017context}
Vedantam, R., Bengio, S., Murphy, K., Parikh, D., Chechik, G.: Context-{Aware}
  {Captions} from {Context}-{Agnostic} {Supervision}. In: 2017 {IEEE}
  {Conference} on {Computer} {Vision} and {Pattern} {Recognition} ({CVPR})
  (2017)

\bibitem{vedantam2015cider}
Vedantam, R., Zitnick, C.L., Parikh, D.: {CIDEr}: {Consensus}-based image
  description evaluation. In: 2015 {IEEE} {Conference} on {Computer} {Vision}
  and {Pattern} {Recognition} ({CVPR}) (2015)

\bibitem{vijayakumar2018diverse}
Vijayakumar, A.K., Cogswell, M., Selvaraju, R.R., Sun, Q., Lee, S., Crandall,
  D., Batra, D.: Diverse {Beam} {Search} for {Improved} {Description} of
  {Complex} {Scenes}. In: Thirty-{Second} {AAAI} {Conference} on {Artificial}
  {Intelligence} (2018)

\bibitem{vijayakumar2016diverse}
Vijayakumar, A.K., Cogswell, M., Selvaraju, R.R., Sun, Q., Lee, S., Crandall,
  D., Batra, D.: Diverse {Beam} {Search}: {Decoding} {Diverse} {Solutions} from
  {Neural} {Sequence} {Models}. arXiv:1610.02424 [cs]  (2018), arXiv:
  1610.02424

\bibitem{vinyals2015show}
Vinyals, O., Toshev, A., Bengio, S., Erhan, D.: Show and tell: {A} neural image
  caption generator. In: 2015 {IEEE} {Conference} on {Computer} {Vision} and
  {Pattern} {Recognition} ({CVPR}) (2015)

\bibitem{vinyals2017show}
Vinyals, O., Toshev, A., Bengio, S., Erhan, D.: Show and {Tell}: {Lessons}
  {Learned} from the 2015 {MSCOCO} {Image} {Captioning} {Challenge} (2017)

\bibitem{wang2019describing}
Wang, Q., Chan, A.B.: Describing {Like} {Humans}: {On} {Diversity} in {Image}
  {Captioning}. In: 2019 {IEEE}/{CVF} {Conference} on {Computer} {Vision} and
  {Pattern} {Recognition} ({CVPR}) (2019)

\bibitem{wu2017diverse}
Wu, B., Jia, F., Liu, W., Ghanem, B.: Diverse {Image} {Annotation}. In: 2017
  {IEEE} {Conference} on {Computer} {Vision} and {Pattern} {Recognition}
  ({CVPR}) (2017)

\bibitem{xu2015show}
Xu, K., Ba, J.L., Kiros, R., Cho, K., Courville, A., Salakhutdinov, R., Zemel,
  R.S., Bengio, Y.: Show, attend and tell: neural image caption generation with
  visual attention. In: Proceedings of the 32nd {International} {Conference} on
  {International} {Conference} on {Machine} {Learning} (ICML) (2015)

\bibitem{AttnGAN}
Xu, T., Zhang, P., Huang, Q., Zhang, H., Gan, Z., Huang, X., He, X.: {AttnGAN}:
  {Fine}-{Grained} {Text} to {Image} {Generation} with {Attentional}
  {Generative} {Adversarial} {Networks}. In: 2018 {IEEE}/{CVF} {Conference} on
  {Computer} {Vision} and {Pattern} {Recognition} (CVPR) (2018)

\bibitem{yao2010co}
Yao, T., Mei, T., Ngo, C.W.: Co-reranking by mutual reinforcement for image
  search. In: Proceedings of the {ACM} {International} {Conference} on {Image}
  and {Video} {Retrieval} (CIVR) (2010)

\bibitem{you_image_2016}
You, Q., Jin, H., Wang, Z., Fang, C., Luo, J.: Image {Captioning} with
  {Semantic} {Attention}. In: 2016 {IEEE} {Conference} on {Computer} {Vision}
  and {Pattern} {Recognition} ({CVPR}) (2016)

\bibitem{zhang2017stackgan}
Zhang, H., Xu, T., Li, H., Zhang, S., Wang, X., Huang, X., Metaxas, D.:
  {StackGAN}: {Text} to {Photo}-{Realistic} {Image} {Synthesis} with {Stacked}
  {Generative} {Adversarial} {Networks}. In: 2017 {IEEE} {International}
  {Conference} on {Computer} {Vision} ({ICCV}) (2017)

\bibitem{zhang2018generating}
Zhang, Y., Galley, M., Gao, J., Gan, Z., Li, X., Brockett, C., Dolan, B.:
  Generating {Informative} and {Diverse} {Conversational} {Responses} via
  {Adversarial} {Information} {Maximization}. In: Advances in {Neural}
  {Information} {Processing} {Systems} (2018)

\end{thebibliography}
